\documentclass[10pt,twocolumn,letterpaper]{article}

\usepackage{cvpr}
\usepackage{times}
\usepackage{graphicx}
\usepackage{amsmath}
\usepackage{amssymb}
\usepackage{subcaption}

\usepackage[ruled,lined]{algorithm2e}
\newlength\myindent
\newcommand\hide[1]{}

\usepackage{gensymb}
\usepackage{booktabs}
\usepackage{cite}

\usepackage[pagebackref=true,breaklinks=true,letterpaper=true,colorlinks,bookmarks=false]{hyperref}

\cvprfinalcopy 

\def\cvprPaperID{0000} 

\ifcvprfinal\fi
\begin{document}
\tracingall
\title{Bottom-up Object Detection by Grouping Extreme and Center Points}

\author{Xingyi Zhou\\
UT Austin\\
{\tt\small zhouxy@cs.utexas.edu}
\and
Jiacheng Zhuo\\
UT Austin\\
{\tt\small jzhuo@cs.utexas.edu}
\and
Philipp Kr\"ahenb\"uhl\\
UT Austin\\
{\tt\small philkr@cs.utexas.edu}
}

\maketitle

\begin{abstract}
With the advent of deep learning, object detection drifted from a bottom-up to a top-down recognition problem.
State of the art algorithms enumerate a near-exhaustive list of object locations and classify each into: object or not.
In this paper, we show that bottom-up approaches still perform competitively.
We detect four extreme points (top-most, left-most, bottom-most, right-most) and one center point of objects using a standard keypoint estimation network.
We group the five keypoints into a bounding box if they are geometrically aligned. 
Object detection is then a purely appearance-based keypoint estimation problem, without region classification or implicit feature learning. 
The proposed method performs on-par with the state-of-the-art region based detection methods, with a bounding box AP of $43.7\%$ on COCO test-dev.
In addition, our estimated extreme points directly span a coarse octagonal mask, with a COCO Mask AP of $18.9\%$, much better than the Mask AP of vanilla bounding boxes.
Extreme point guided segmentation further improves this to $34.6\%$ Mask AP.
\end{abstract}


\section{Introduction}

\begin{figure}[t]
\begin{subfigure}[b]{0.22\textwidth}
\begin{center}
   \includegraphics[width=0.95\linewidth]{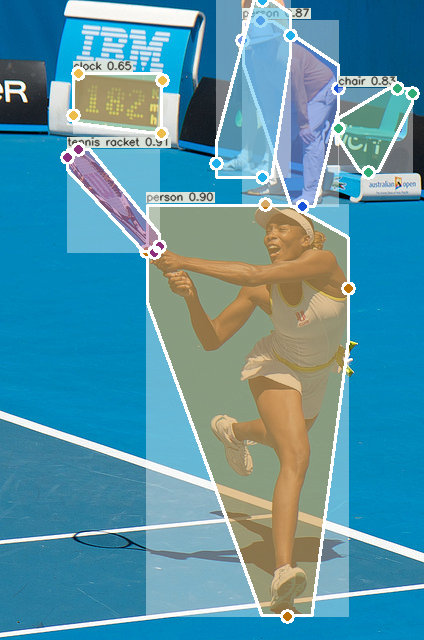}
\end{center}
\end{subfigure} \hfill
\begin{subfigure}[b]{0.248\textwidth}
\begin{center}
   \includegraphics[width=0.95\linewidth]{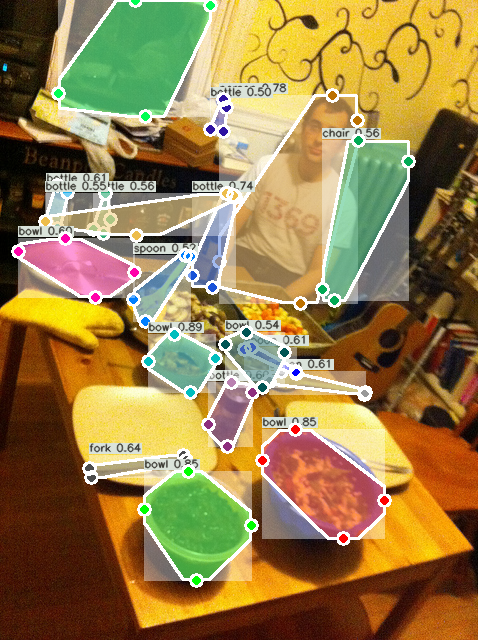}
\end{center}
\end{subfigure}
   \caption{We propose to detect objects by finding their extreme points. They directly form a bounding box , but also give a much tighter octagonal approximation of the object.}
\label{fig:oct}
\end{figure}

Top-down approaches have dominated object detection for years. 
Prevalent detectors convert object detection into rectangular region classification, by either explicitly cropping the region~\cite{girshick2014rich} or region feature~\cite{girshick2015fast,ren2015faster}  (two-stage object detection) or implicitly setting fix-sized anchors for region proxies~\cite{lin2018focal,redmon2016you,liu2016ssd} (one-stage object detection). 
However, top-down detection is not without limits.
A rectangular bounding box is not a natural object representation.
Most objects are not axis-aligned boxes, and fitting them inside a box includes many distracting background pixels (Figure. ~\ref{fig:oct}).
In addition, top-down object detectors enumerate a large number of possible box locations without truly understanding the compositional visual grammars~\cite{felzenszwalb2010object,girshick2011object} of objects themselves.
This is computationally expensive.
Finally, boxes are a bad proxy for the object themselves.
They convey little detailed object information, e.g., object shape and pose.

\begin{figure}[t]
\begin{center}
\includegraphics[width=0.95\linewidth]{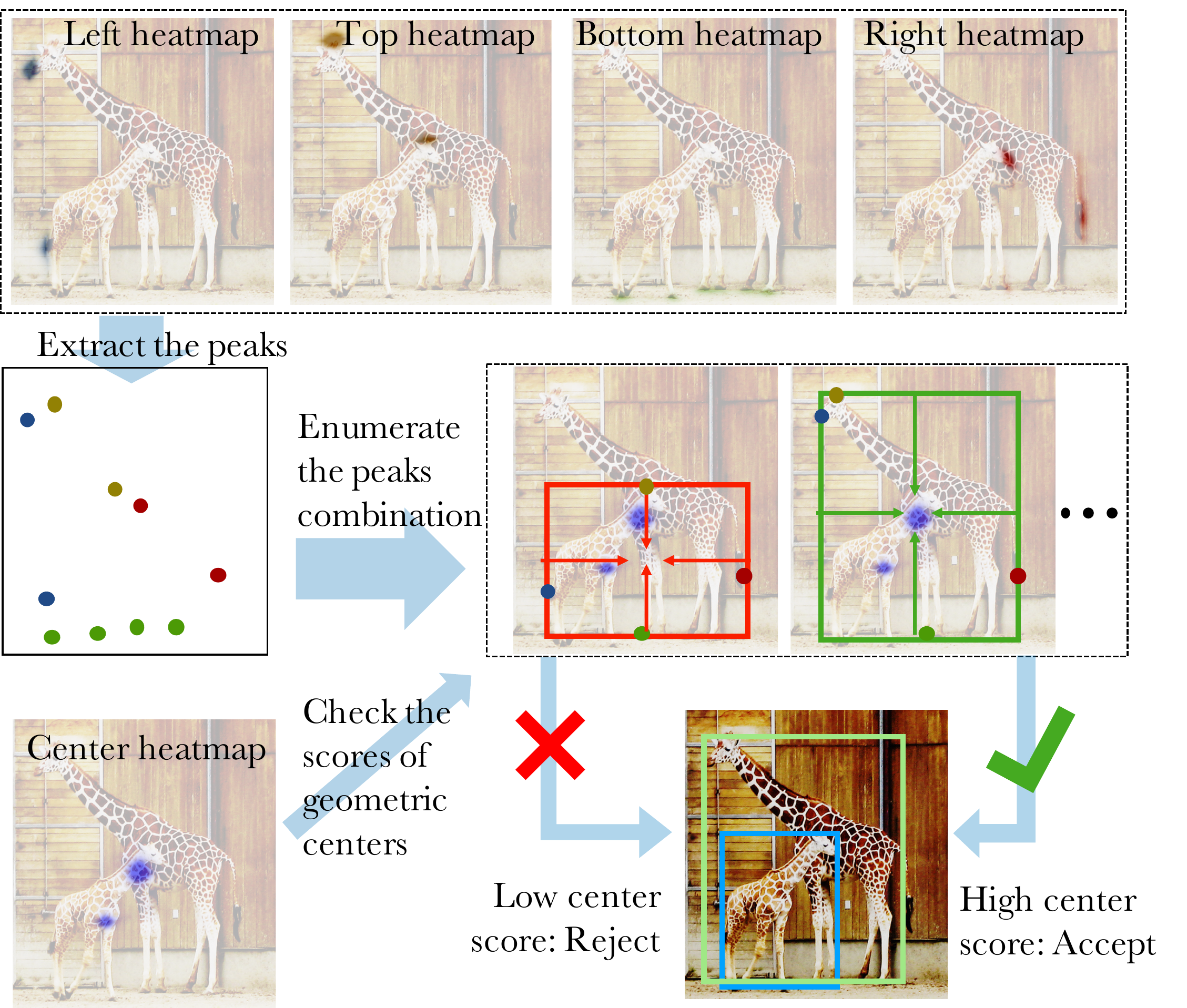}
\vspace{-1em}
\end{center}
   \caption{Illustration of our object detection method. Our network predicts four extreme point heatmaps (Top. We shown the heatmap overlaid on the input image) and one center heatmap (Bottom row left) for each category. We enumerate the combinations of the peaks (Middle left) of four extreme point heatmaps and compute the geometric center of the composed bounding box (Middle right). A bounding box is produced if and only if its geometric center has a high response in the center heatmap (Bottom right).}
\label{fig:group}
\end{figure}

In this paper, we propose ExtremeNet, a bottom-up object detection framework that detects four extreme points (top-most, left-most, bottom-most, right-most) of an object. 
We use a state-of-the-art keypoint estimation framework~\cite{newell2016stacked,Chen_2018_CVPR,xiao2018simple,newell2017associative,cao2017realtime} to find extreme points, by predicting four multi-peak heatmaps for each object category.
In addition, we use one heatmap per category predicting the object center, as the average of two bounding box edges in both the $x$ and $y$ dimension.
We group extreme points into objects with a purely geometry-based approach.
We group four extreme points, one from each map, if and only if their geometric center is predicted in the center heatmap with a score higher than a pre-defined threshold.
We enumerate all $O(n^4)$ combinations of extreme point prediction, and select the valid ones.
The number of extreme point prediction $n$ is usually quite small, for COCO~\cite{lin2014microsoft} $n \le 40$, and a brute force algorithm implemented on GPU is sufficient.
Figure~\ref{fig:group} shows an overview of the proposed method.

We are not the first to use deep keypoint prediction for object detection.
CornerNet~\cite{Law_2018_ECCV} predicts two opposing corners of a bounding box.
They group corner points into bounding boxes using an associative embedding feature~\cite{newell2017associative}. 
Our approach differs in two key aspects: keypoint definition and grouping. 
A corner is another form of bounding box, and suffers many of the issues top-down detection suffers from.
A corner often lies outside an object, without strong appearance features.
Extreme points, on the other hand, lie on objects, are visually distinguishable, and have consistent local appearance features.
For example, the top-most point of human is often the head, and the bottom-most point of a car or airplane will be a wheel.
This makes the extreme point detection easier.
The second difference to CornerNet is the geometric grouping.
Our detection framework is fully appearance-based, without any implicit feature learning.
In our experiments, the appearance-based grouping works significantly better.

Our idea is motivated by Papadopoulos et al.~\cite{papadopoulos2017extreme}, who proposed to annotate bounding boxes by clicking the four extreme points.
This annotation is roughly four times faster to collect and provides richer information than bounding boxes.
Extreme points also have a close connection to object masks.
Directly connecting the inflated extreme points offers a more fine-grained object mask than the bounding box. 
In our experiment, we show that fitting a simple octagon to the extreme points yields a good object mask estimation.
Our method can be further combined with Deep Extreme Cut (DEXTR)~\cite{Man+18}, which turns extreme point annotations into a segmentation mask for the indicated object.
Directly feeding our extreme point predictions as guidance to DEXTR~\cite{Man+18} leads to close to state-of-the-art instance segmentation results.

Our proposed method achieves a bounding box AP of $43.7\%$ on COCO test-dev, out-performing all reported one-stage object detectors~\cite{lin2018focal,zhang2018single,redmon2018yolov3,Law_2018_ECCV} and on-par with sophisticated two-stage detectors.
A Pascal VOC~\cite{pascal-voc-2012,BharathICCV2011} pre-trained DEXTR~\cite{Man+18} model yields a Mask AP of $34.6\%$, without using any COCO mask annotations. 
Code is available at \url{https://github.com/xingyizhou/ExtremeNet}. 

\section{Related Work}

\textbf{Two-stage object detectors} 
Region-CNN family~\cite{girshick2014rich,he2014spatial,girshick2015fast,ren2015faster,he2017mask} considers object detection as two sequential problems: first propose a (large) set of \emph{category-agnostic} bounding box candidates, crop them, and use an image classification module to classify the cropped region or region feature.
R-CNN~\cite{girshick2014rich} uses selective search ~\cite{uijlings2013selective} to generate region proposals and feeds them to an ImageNet classification network. 
SPP~\cite{he2014spatial} and Fast RCNN~\cite{girshick2015fast} first feed an image through a convolutional network and crop an intermediate feature map to reduce computation.
Faster RCNN~\cite{ren2015faster} further replaces region proposals~\cite{uijlings2013selective} with a Region Proposal Network.
The detection-by-classification idea is intuitive and keeps the best performance so far~\cite{dai2016r,Dai_2017_ICCV,peng2017megdet,liu2018path,lin2017feature,zhu2017couplenet,huang2017speed,tychsen2017improving,tychsen2017denet,Jiang_2018_ECCV}.

Our method does not require region proposal or region classification. 
We argue that a \emph{region} is not a necessary component in object detection. 
Representing an object by four extreme points is also effective and provides as much information as bounding boxes.

\begin{figure*}[t]
\begin{center}
\includegraphics[width=0.95\linewidth]{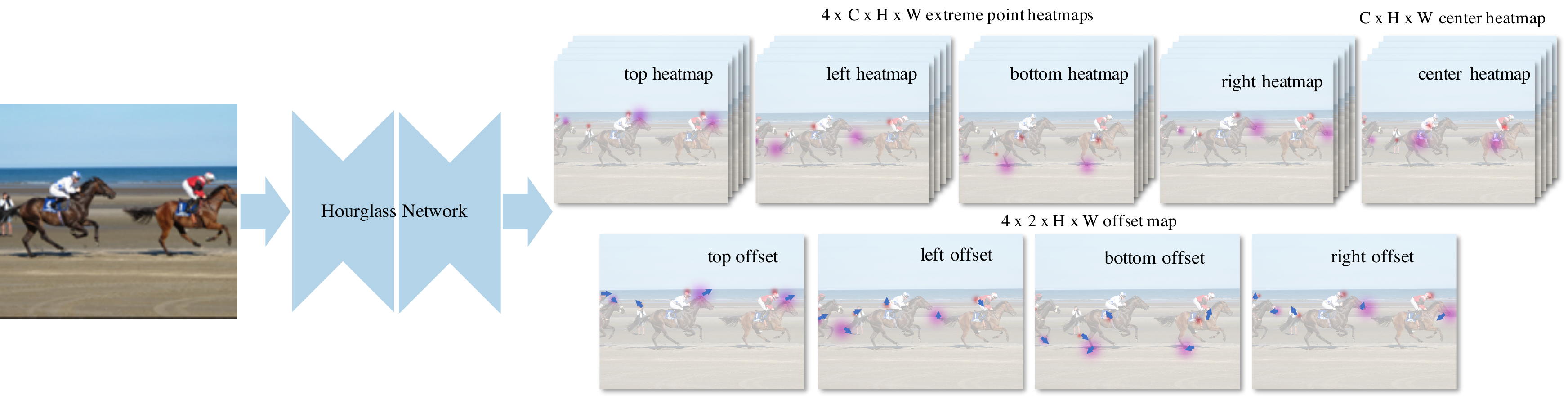}
\vspace{-1em}
\end{center}
   \caption{Illustration of our framework. Our network takes an image as input and produces four C-channel heatmaps, one C-channel heatmap, and four 2-channel category-agnostic offset map. The heatmaps are trained by weighted pixel-wise logistic regression, where the weight is used to reduce false-positive penalty near the ground truth location. And the offset map is trained with Smooth L1 loss applied at ground truth peak locations.}
\label{fig:framework}
\end{figure*}

\textbf{One-stage object detector}
One-stage object detectors~\cite{lin2018focal,redmon2016you,liu2016ssd,redmon2017yolo9000,Law_2018_ECCV,wang2017point,shen2017dsod} do not have a region cropping module. 
They can be considered as \emph{category-specific} region or anchor proposal networks and directly assign a class label to each positive anchor. 
SSD~\cite{liu2016ssd,fu2017dssd} uses different scale anchors in different network layers. 
YOLOv2~\cite{redmon2017yolo9000} learns anchor shape priors.
RetinaNet~\cite{lin2018focal} proposes a focal loss to balance the training contribution between positive and negative anchors. 
RefineDet~\cite{zhang2018single} learns to early reject negative anchors.
Well-designed single-stage object detectors achieve very close performance with two-stage ones at higher efficiency. 

Our method falls in the one-stage detector category. 
However, instead of setting anchors in an $O(h^2w^2)$ space, we detects five individual parts (four extreme points and one center) of a bounding box in $O(hw)$ space. 
Instead of setting default scales or aspect-ratios as anchors at each pixel location, we only predict the probability for that location being a keypoint. 
Our center map can also be seen as a scale and aspect ratio agnostic region proposal network without bounding box regression. 

\textbf{Deformable Part Model}
As a bottom-up object detection method, our idea of grouping center and extreme points is related to Deformable Part Model~\cite{felzenszwalb2010object}. 
Our center point detector functions similarly with the root filter in DPM~\cite{felzenszwalb2010object}, and our
four extreme points can be considered as a universal part decomposition for all categories. 
Instead of learning the part configuration, our predicted center and four extreme points have a \emph{fixed} geometry structure.
And we use a state-of-the-art keypoint detection network instead of low-level image filters for part detection. 

\textbf{Grouping in bottom-up human pose estimation}
Determining which keypoints are from the same person is an important component in bottom-up multi-person pose estimation.
There are multiple solutions:
Newell \etal ~\cite{newell2017associative} proposes to learn an associative feature for each keypoint, which is trained using an embedding loss.
Cao \etal ~\cite{cao2017realtime} learns an affinity field which resembles the edge between connected keypoints.
Papandreous \etal ~\cite{papandreou2018personlab} learns the displacement to the parent joint on the human skeleton tree, as a 2-d feature for each keypoint. 
Nie \etal ~\cite{nie2018ppn} also learn a feature as the offset with respect to the object center. 

In contrast to all the above methods, our center grouping is pure appearance-based and is easy to learn, by exploiting the geometric structure of extreme points and their center.

\textbf{Implicit keypoint detection}
Prevalent keypoint detection methods work on well-defined semantic keypoints, e.g., human joints. 
StarMap~\cite{zhou2018starmap} mixes all types of keypoints using a single heatmap for general keypoint detection. 
Our extreme and center points are a kind of such general implicit keypoints, but with more explicit geometry property.

\section{Preliminaries}

\paragraph{Extreme and center points}

Let $(x^{(tl)}, y^{(tl)}, x^{(br)}, y^{(br)})$ denote the four sides of a bounding box.
To annotate a bounding box, a user commonly clicks on the top-left $(x^{(tl)}, y^{(tl)})$ and bottom-right $(x^{(br)}, y^{(br)})$ corners.
As both points regularly lie outside an object, these clicks are often inaccuracy and need to be adjusted a few times.
The whole process takes $34.5$ seconds on average~\cite{su2012crowdsourcing}. 
Papadopoulos et al.~\cite{papadopoulos2017extreme} propose to annotate the bounding box by clicking the four extreme points $(x^{(t)}, y^{(t)})$, $(x^{(l)}, y^{(l)})$, $(x^{(b)}, y^{(b)})$, $(x^{(r)}, y^{(r)})$, where the box is $(x^{(l)}, y^{(t)}, x^{(r)}, y^{(b)})$.
An extreme point is a point $(x^{(a)}, y^{(a)})$ such that no other point $(x,y)$ on the object lies further along one of the four cardinal directions $a$: top, bottom, left, right.
Extreme click annotation time is 7.2 seconds on average~\cite{papadopoulos2017extreme}.
The resulting annotation is on-par with the more time-consuming box annotation.
Here, we use the extreme click annotations directly and bypass the bounding box.
We additionally use the center point of each object as $(\frac{x^{(l)} + x^{(r)}}{2}, \frac{y^{(t)} + y^{(b)}}{2})$. 

\paragraph{Keypoint detection}

Keypoint estimation, e.g., human joint estimation~\cite{newell2017associative,cao2017realtime,Chen_2018_CVPR,xiao2018simple,he2017mask} or chair corner point estimation~\cite{zhou2018starmap,pavlakos20176}, commonly uses a fully convolutional encoder-decoder network to predict a multi-channel heatmap for each type of keypoint (e.g., one heatmap for human head, another heatmap for human wrist).
The network is trained in a fully supervised way, with either an L2 loss to a rendered Gaussian map~\cite{newell2017associative,cao2017realtime,Chen_2018_CVPR,xiao2018simple} or with a per-pixel logistic regression loss~\cite{papandreou2017towards,papandreou2018personlab,Law_2018_ECCV}.
State-of-the-art keypoint estimation networks, e.g., 104-layer HourglassNet~\cite{newell2016stacked,Law_2018_ECCV}, are trained in a fully convolutional manner.
They regress to a heatmap $\hat{Y} \in (0, 1)^{H \times W}$ of width $W$ and height $H$ for each output channel.
The training is guided by a multi-peak Gaussian heatmap $Y \in (0, 1)^{H \times W}$, where each keypoint defines the mean of a Gaussian Kernel.
The standard deviation is either fixed, or set proportional to the object size~\cite{Law_2018_ECCV}.
The Gaussian heatmap serves as the regression target in the L2 loss case or as the weight map to reduce penalty near a positive location in the logistic regression case~\cite{Law_2018_ECCV}. 

\paragraph{CornerNet}
CornerNet~\cite{Law_2018_ECCV} uses keypoint estimation with an HourglassNet~\cite{newell2016stacked} as an object detector.
They predict two sets of heatmaps for the opposing corners of the box.
In order to balance the positive and negative locations they use a modified focal loss~\cite{lin2018focal} for training:
\begin{equation}
    \!\!L_{det}\!=\!-\!\frac{1}{N}\!\sum^{H}_{i=1}\!\sum^{W}_{j=1}\!
    \begin{matrix}
        (1 - \hat{Y}_{ij})^{\alpha} 
        \log(\hat{Y}_{ij}) &\hspace{-3mm} \text{if}\ Y_{ij}\!=\!1\\ 
        \!(\!1\!-\!Y_{ij}\!)^{\beta} 
        (\hat{Y}_{ij}\!)^{\alpha}
        \log(\!1\!-\!\hat{Y}_{ij}\!)\!\! &\hspace{-3mm} \text{o.w.}
    \end{matrix},
    \label{eq:det_loss}
\end{equation}
where $\alpha$ and $\beta$ are hyper-parameters and fixed to $\alpha=2$ and $\beta=4$ during training.
$N$ is the number of objects in the image.

For sub-pixel accuracy of extreme points, CornerNet additionally regresses to category-agnostic keypoint offset $\Delta^{(a)}$ for each corner.
This regression recovers part of the information lost in the down-sampling of the hourglass network.
The offset map is trained with smooth L1 Loss~\cite{girshick2015fast} $SL_1$ on ground truth extreme point locations:
\begin{equation}
    L_{off} = \frac{1}{N}\sum_{k=1}^{N} SL_1(\Delta^{(a)}, \vec x/s - \lfloor \vec x/s \rfloor ),
\end{equation}
where $s$ is the down-sampling factor ($s=4$ for HourglassNet), $\vec x$ is the coordinate of the keypoint.

CornerNet then groups opposing corners into detection using an associative embedding~\cite{newell2017associative}.
Our extreme point estimation uses the CornerNet architecture and loss, but not the associative embedding.

\paragraph{Deep Extreme Cut}
Deep Extreme Cut (DEXTR)~\cite{Man+18} is an extreme point guided image segmentation method. 
It takes four extreme points and the cropped image region surrounding the bounding box spanned by the extreme points as input.
From this it produces a \emph{category-agnostic} foreground segmentation of the indicated object using the semantic segmentation network of Chen \etal~\cite{chen2018deeplab}.
The network learns to generate the segmentation mask that matches the input extreme point. 

\section{ExtremeNet for Object detection}

ExtremeNet uses an HourglassNet~\cite{newell2016stacked} to detect five keypoints per class (four extreme points, and one center).
We follow the training setup, loss and offset prediction of CornerNet~\cite{Law_2018_ECCV}.
The offset prediction is category-agnostic, but extreme-point specific.
There is no offset prediction for the center map.
The output of our network is thus $5 \times C$ heatmaps and $4 \times 2$ offset maps, where $C$ is the number of classes ($C = 80$ for MS COCO~\cite{lin2014microsoft}).
Figure~\ref{fig:framework} shows an overview.
Once the extreme points are extracted, we group them into detections in a purely geometric manner.

\subsection{Center Grouping}
\label{sec:grouping}
\begin{algorithm}[t]\footnotesize
	\caption{\footnotesize Center Grouping}
	\label{alg:size_balance}
	\SetKwInOut{Input}{Input} \SetKwInOut{Output}{Output}
		 \Input{Center and Extremepoint heatmaps of an image for one category: $\hat Y^{(c)}$, $\hat Y^{(t)}$, $\hat Y^{(l)}$, $\hat Y^{(b)}$, $\hat Y^{(r)}$  $\in (0, 1)^{H \times W}$ \\ 
		 Center and peak selection thresholds: $\tau_c$ and $\tau_p$}
		 \Output{Bounding box with score}
		 // Convert heatmaps into coordinates of keypoints. \\
		 // $\mathcal{T}, \mathcal{L}, \mathcal{B}, \mathcal{R}$ are sets of points. \\
		 $\mathcal{T} \leftarrow \textrm{ExtractPeak}(\hat Y^{(t)}, \tau_p)$ \\
		 $\mathcal{L} \leftarrow \textrm{ExtractPeak}(\hat Y^{(l)}, \tau_p)$ \\
		 $\mathcal{B} \leftarrow \textrm{ExtractPeak}(\hat Y^{(b)}, \tau_p)$ \\
		 $\mathcal{R} \leftarrow \textrm{ExtractPeak}(\hat Y^{(r)}, \tau_p)$ \\
		 \For{$t \in \mathcal{T}$, $l \in \mathcal{L}$, $b \in \mathcal{B}$, $r \in \mathcal{R}$} {
		    // If the bounding box is valid\\
		    \If{$t_y \le l_y, r_y \le b_y$ \textbf{and} $l_x \le t_x, b_x \le r_x$} {
                // compute geometry center \\
                $c_x \leftarrow (l_x + r_x) / 2$ \\
                $c_y \leftarrow (t_y + b_y) / 2$ \\
                // If the center is detected \\
                \If {$\hat Y^{(c)}_{c_x, c_y} \ge \tau_c$} {
                    Add Bounding box $(l_x, t_y, r_x, b_y)$ with score $(\hat Y^{(t)}_{t_x, t_y} + \hat Y^{(l)}_{l_x, l_y} + \hat Y^{(b)}_{b_x, b_y} + \hat Y^{(r)}_{r_x, r_y} + \hat Y^{(c)}_{c_x, c_y})/5$.
                }
            }
		 }
	\label{alg:grouping}

\end{algorithm}

Extreme points lie on different sides of an object.
This complicates grouping.
For example, an associative embedding~\cite{newell2017associative} might not have a global enough view to group these keypoints.
Here, we take a different approach that exploits the spread out nature of extreme points.

The input to our grouping algorithm is five heatmaps per class: one center heatmap $
\hat Y^{(c)} \in (0, 1)^{H \times W}$ and four extreme heatmaps $\hat Y^{(t)}$, $\hat Y^{(l)}$, $\hat Y^{(b)}$, $\hat Y^{(r)} \in (0, 1)^{H \times W}$ for the top, left, bottom, right, respectively.
Given a heatmap, we extract the corresponding keypoints by detecting all peaks.
A peak is any pixel location with a value greater than $\tau_p$, that is locally maximal in a $3 \times 3$ window surrounding the pixel.
We name this procedure as \textit{ExtrectPeak}.

Given four extreme points $t$, $b$, $r$, $l$ extracted from heatmaps $\hat Y^{(t)}$, $\hat Y^{(l)}$, $\hat Y^{(b)}$, $\hat Y^{(r)}$, we compute their geometric center $c = (\frac{l_x+t_x}{2}, \frac{t_y+b_y}{2})$.
If this center is predicted with a high response in the center map $\hat Y^{(c)}$, we commit the extreme points as a valid detection: $\hat Y^{(c)}_{c_x,c_y} \ge \tau_c$ for a threshold $\tau_c$.
We then enumerate over all quadruples of keypoints $t$, $b$, $r$, $l$ in a brute force manner.
We extract detections for each class independently.
Algorithm \ref{alg:grouping} summarizes this procedure.
We set $\tau_p = 0.1$ and $\tau_c = 0.1$ in all experiments.

This brute force grouping algorithm has a runtime of $O(n^4)$, where $n$ is the number of extracted extreme points for each cardinal direction.
Supplementary material presents a $O(n^2)$ algorithm that is faster on paper.
However, then it is harder to accelerate on a GPU and slower in practice for the MS COCO dataset, where $n \le 40$.

\subsection{Ghost box suppression}
\label{sec:ghost}
Center grouping may give a high-confidence false-positive detection for three equally spaced colinear objects of the same size.
The center object has two choices here, commit to the correct small box, or predict a much larger box containing the extreme points of its neighbors.
We call these false-positive detections ``ghost'' boxes.
As we'll show in our experiments, these ghost boxes are infrequent, but nonetheless a consistent error mode of our grouping.

We present a simple post-processing step to remove ghost boxes.
By definition a ghost box contains many other smaller detections.
To discourage ghost boxes, we use a form of soft non-maxima suppression~\cite{bodla2017soft}.
If the sum of scores of all boxes \emph{contained} in a certain bounding box exceeds $3$ times of the score of itself,  we divide its score by $2$.
This non-maxima suppression is similar to the standard overlap-based non-maxima suppression, but penalizes potential ghost boxes instead of multiple overlapping boxes.

\subsection{Edge aggregation}

{
\begin{figure}[t]
\begin{subfigure}[b]{0.235\textwidth}
\begin{center}
   \includegraphics[width=0.95\linewidth]{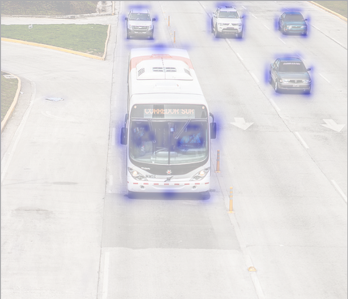}
   \caption{Original heatmap.}
\end{center}
\end{subfigure} \hfill
\begin{subfigure}[b]{0.235\textwidth}
\begin{center}
   \includegraphics[width=0.95\linewidth]{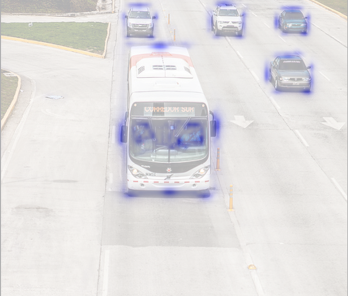}
   \caption{After edge aggregation.}
\end{center}
\end{subfigure}
   \caption{Illustration of the purpose of edge aggregation. In the case of multiple points being the extreme point on one edge, our model predicts a segment of low confident responses (a). Edge aggregation enhances the confidence of the middle pixel (b).}
\label{fig:aggregation}
\end{figure}
}

\label{sec:edgeAggr}
Extreme points are not always uniquely defined.
If vertical or horizontal edges of an object form the extreme points (e.g., the top of a car) any point along that edge might be considered an extreme point.
As a result, our network produces a weak response along any aligned edges of the object, instead of a single strong peak response.
This weak response has two issues: First, the weaker response might be below our peak selection threshold $\tau_p$, and we will miss the extreme point entirely. Second, even if we detect the keypoint, its score will be lower than a slightly rotated object with a strong peak response.

We use edge aggregation to address this issue.
For each extreme point, extracted as a local maximum, we aggregate its score in either the vertical direction, for left and right extreme points, or the horizontal direction, for top and bottom keypoints.
We aggregate all monotonically decreasing scores, and stop the aggregation at a local minimum along the aggregation direction.
Specifically, let $m$ be an extreme point and $N^{(m)}_i = \hat Y_{m_x+i, m_y}$ be the vertical or horizontal line segment at that point.
Let $i_0 < 0$ and $0 < i_1$ be the two closest local minima $N^{(m)}_{i_0-1} > N^{(m)}_{i_0}$ and $N^{(m)}_{i_1} < N^{(m)}_{i_1+1}$.
Edge aggregation updates the keypoint score as $\tilde Y_{m} = \hat Y_{m} + \lambda_{aggr} \sum_{i=i_0}^{i_1} N^{(m)}_i$, where $\lambda_{aggr}$ is the aggregation weight.
In our experiments, we set $\lambda_{aggr}=0.1$.
See Fig. ~\ref{fig:aggregation} for en example.

\subsection{Extreme Instance Segmentation}

\label{sec:ins}

Extreme points carry considerable more information about an object, than a simple bounding box, with at least twice as many annotated values (8 vs 4).
We propose a simple method to approximate the object mask using extreme points by creating an octagon whose edges are centered on the extreme points.
Specifically, for an extreme point, we extend it in both directions on its corresponding edge to a segment of $1/4$ of the entire edge length.
The segment is truncated when it meets a corner. 
We then connect the end points of the four segments to form the octagon. 
See Figure ~\ref{fig:oct} for an example. 

To further refine the bounding box segmentation, we use Deep Extreme Cut (DEXTR)~\cite{Man+18}, a deep network trained to convert the manually provided extreme points into instance segmentation mask.
In this work, we simply replace the manual input of DEXTR~\cite{Man+18} with our extreme point prediction, to perform a 2-stage instance segmentation.
Specifically, for each of our predicted bounding box \hide{with a score $> 0.2$}, we crop the bounding box region, render a Gaussian map with our predicted extreme point, and then feed the concatenated image to the pre-trained DEXTR model. 
DEXTR~\cite{Man+18} is class-agnostic, thus we directly use the detected class and score of ExtremeNet.
No further post-processing is used. 

\section{Experiments}

We evaluate our method on the popular MS COCO dataset~\cite{lin2014microsoft}. 
COCO contains rich bounding box and instance segmentation annotations for 80 categories. 
We train on the train2017 split, which contains 118k images and 860k annotated objects.
We perform all ablation studies on val2017 split, with 5k images and 36k objects, and compare to prior work on the test-dev split with contains 20k images
The main evaluation metric is average precision over a dense set of fixed recall threshold
We show average precision at IOU threshold $0.5$ ($AP_{50}$), $0.75$ ($AP_{75}$), and averaged over all thresholds between $0.5$ and $1$ ($AP$).
We also report AP for small, median and large objects ($AP_S$, $AP_M$, $AP_L$).
The test evaluation is done on the official evaluation server. 
Qualitative results are shown in Table.~\ref{table:demo} and can be found more in the supplementary material.

\subsection{Extreme point annotations}

There are no direct extreme point annotation in the COCO~\cite{lin2014microsoft}.
However, there are complete annotations for object segmentation masks.
We thus find extreme points as extrema in the polygonal mask annotations.
In cases where an edge is parallel to an axis or within a $3^{\degree}$ angle, we place the extreme point at the center of the edge.
Although our training data is derived from the more expensive segmentation annotation, the extreme point data itself is $4 \times$ cheaper to collect than the standard bounding box~\cite{papadopoulos2017extreme}.

\subsection{Training details}

Our implementation is based on the public implementation of CornerNet~\cite{Law_2018_ECCV}.
We strictly follow CornerNets hyper-parameters:
we set the input resolution to $511 \times 511$ and output resolution to $128 \times 128$. 
Data augmentation consists of flipping, random scaling between $0.6$ and $1.3$, random cropping, and random color jittering.
The network is optimized with Adam~\cite{kingma2014adam} with learning rate $2.5e-4$. 
CornerNet~\cite{Law_2018_ECCV} was originally trained on 10 GPUs for 500k iterations, and an equivalent of over $140$ GPU days.
Due to limited GPU resources, the self-comparison experiments (Table. ~\ref{table:Ablation}) are finetuned from a pre-trained CornerNet model with randomly initialized head layers on 5 GPUs for $250k$ iterations with a batch size of 24. 
Learning rate is dropped $10 \times$ at the $200k$ iteration.
The state-of-the-art comparison experiment is trained from scratch on 5 GPUs for $500k$ iterations with learning rate dropped at the $450k$ iteration.

\subsection{Testing details}
For each input image, our network produces four $C$-channel heatmaps for extreme points, one $C$-channel heatmap for center points, and four $2$-channel offset maps. 
We apply edge aggregation (Section.~\ref{sec:edgeAggr}) to each extreme point heatmap, and multiply the center heatmap by $2$ to correct for the overall scale change. 
We then apply the center grouping algorithm (Section.~\ref{sec:grouping}) to the heatmaps.
At most $40$ top points are extracted in \emph{ExtrectPeak} to keep the enumerating efficiency. 
The predicted bounding box coordinates are refined by adding an offset at the corresponding location of offsetmaps. 

Following CornerNet~\cite{Law_2018_ECCV}, we keep the original image resolution instead of resizing it to a fixed size.  
We use flip augmentation for testing.
In our main comparison, we use additional $5 \times$ multi-scale (0.5, 0.75, 1, 1.25, 1.5) augmentation. 
Finally, Soft-NMS~\cite{bodla2017soft} filters all augmented detection results. 
Testing on one image takes 322ms (3.1FPS), with 168ms on network forwarding, 130ms on decoding and rest time on image pre- and post-processing (NMS). 

\subsection{Ablation studies}

\setlength{\tabcolsep}{2pt}
\begin{table}
\begin{center}
\begin{tabular}{lcccccc}
\toprule
 & $AP$  & $AP_{50}$ & $AP_{75}$ & $AP_{S}$ & $AP_{M}$ & $AP_{L}$ \\
\midrule
 & 40.3 & 55.1 & 43.7 & 21.6 & 44.0 & 56.1\\
w/ multi-scale testing & 43.3 & 59.6 & 46.8 & 25.7 & 46.6 & 59.4\\
\midrule
w/o Center grouping & 38.2 & 53.8 & 40.4 & 20.6 & 41.5 & 52.9 \\
w/o Edge aggregation & 39.6 & 54.7 & 43.0 & 22.0 & 43.0 & 54.1 \\
w/o Ghost removal & 40.0 & 54.7 & 43.3 & 21.6 & 44.2 & 54.1\\
\midrule
w/ gt center & 48.6 & 62.1 & 53.9 & 26.3 & 53.7 & 66.7 \\
w/ gt extreme & 56.3 & 67.2 & 60.0 & 40.9 & 62.0s & 64.0\\
w/ gt extreme + center & 79.8 & 94.5 & 86.2 & 65.5 & 88.7 & 95.5\\
w/ gt ex. + ct. + offset & 86.0 & 94.0 & 91.3 & 73.4 & 95.7 & 98.4\\
\bottomrule
\end{tabular}
\caption{Ablation study and error analysis on COCO val2017. We show AP($\%$) after removing each component or replacing it with its ground truth.}
\label{table:Ablation}
\label{table:Error}
\end{center}
\end{table}
\setlength{\tabcolsep}{1.4pt}

\setlength{\tabcolsep}{2pt}
\begin{table*}
\begin{center}
\begin{tabular}{l@{\ \ }c@{\ \ \ \ \ }c@{\ \ \ \ \ }c@{\ }c@{\ }c@{\ \ \ \ \ }c@{\ }c@{\ }c}
\toprule
 & Backbone & Input resolution & $AP$ & $AP_{50}$ & $AP_{75}$ & $AP_{S}$ & $AP_{M}$ & $AP_{L}$ \\
\midrule
        \multicolumn{9}{l}{\textbf{Two-stage detectors}} \\
        Faster R-CNN w/ FPN~\cite{lin2017feature} & ResNet-101 & $1000 \times 600$ & 36.2 &
        59.1 & 39.0 & 18.2 & 39.0 & 48.2\\ 
        Deformable-CNN~\cite{Dai_2017_ICCV} & Inception-ResNet & $1000 \times 600$ & 37.5 &
        58.0 & - & 19.4 & 40.1 & 52.5 \\
        Deep Regionlets~\cite{Xu_2018_ECCV} & ResNet-101 & $1000 \times 600$ & 39.3 & 59.8 & - & 21.7
        & 43.7 & 50.9\\
        Mask R-CNN~\cite{he2017mask} & ResNeXt-101  & $1333 \times 800$ & 39.8 & 62.3 & 43.4 &
        22.1 & 43.2 & 51.2 \\
        LH R-CNN~\cite{li2017light} & ResNet-101 & $1000 \times 600$ & 41.5 & - & - & 25.2 &
        45.3 & 53.1\\
        Cascade R-CNN~\cite{cai2017cascade} & ResNet-101 & $1333 \times 800$ & 42.8 & 62.1 &
        46.3 & 23.7 & 45.5 & 55.2\\
        D-RFCN + SNIP~\cite{singh2018analysis} & DPN-98  &  $1333 \times 800$
        & 45.7 & 67.3 & 51.1 & 29.3 & 48.8 & 57.1 \\
        PANet~\cite{liu2018path} & ResNeXt-101 & $1000 \times 600$
        & 47.4 & 67.2 & 51.8 & 30.1 & 51.7 & 60.0 \\ 
        \midrule
        \multicolumn{9}{l}{\textbf{One-stage detectors}} \\
        YOLOv2~\cite{redmon2017yolo9000} & DarkNet-19  & $544 \times 544$& 21.6 & 44.0 & 19.2
        & 5.0 & 22.4 & 35.5 \\
        YOLOv3~\cite{redmon2018yolov3} & DarkNet-53 & $608 \times 608$ & 33.0 & 57.9 & 34.4
        & 18.3 & 35.4 & 41.9 \\
        SSD~\cite{liu2016ssd} & ResNet-101 & $513 \times 513$ & 31.2 & 50.4 & 33.3 & 10.2 &
        34.5 & 49.8 \\
        DSSD~\cite{fu2017dssd} & ResNet-101  & $513 \times 513$& 33.2 & 53.3 & 35.2 & 13.0
        & 35.4 & 51.1\\
        RetinaNet~\cite{lin2018focal} & ResNet-101  & $1333 \times 800$& 39.1 & 59.1 & 42.3
        & 21.8 & 42.7 & 50.2 \\
        RefineDet (SS)~\cite{zhang2018single} & ResNet-101 & $512 \times 512$ &
        36.4 & 57.5 & 39.5 & 16.6 & 39.9 & 51.4 \\
        RefineDet (MS)~\cite{zhang2018single} & ResNet-101 & $512 \times 512$ &
        41.8 & 62.9 & 45.7 & 25.6 & 45.1 & 54.1\\
        CornerNet (SS)~\cite{Law_2018_ECCV} & Hourglass-104 & $511 \times 511$ & 40.5 & 56.5 & 43.1 &
        19.4 & 42.7 & 53.9 \\
        CornerNet (MS)~\cite{Law_2018_ECCV} & Hourglass-104 & $511 \times 511$ & 42.1 & 57.8 & 45.3 &
        20.8 & 44.8 & 56.7\\
ExtremeNet (SS)  & Hourglass-104 & $511 \times 511$ & 40.2 & 55.5 & 43.2 & 20.4 & 43.2 & 53.1 \\
ExtremeNet (MS)  & Hourglass-104 & $511 \times 511$ & \emph{43.7} & 60.5 & 47.0 & 24.1 & 46.9 & 57.6 \\
\bottomrule
\end{tabular}
\caption{State-of-the-art comparison on COCO test-dev. SS/ MS are short for single-scale/ multi-scale tesing, respectively. It shows that our ExtremeNet in on-par with state-of-the-art region-based object detectors.}
\label{table:main}
\end{center}
\end{table*}
\setlength{\tabcolsep}{1.4pt}

\setlength{\tabcolsep}{2pt}
\begin{table}[b]
\vspace{-1.5em}
\begin{center}
\begin{tabular}{lcccccccc}
\toprule
 & $AP$  & $AP_{50}$ & $AP_{75}$ & $AP_{S}$ & $AP_{M}$ & $AP_{L}$\\
\midrule
BBox  & 12.1 & 34.9 & 6.2 & 8.2 & 12.7 & 16.9\\
Ours octagon& 18.9 & 44.5 & 13.7 & 10.4 & 20.4 & 28.3 \\
Ours+DEXTR~\cite{Man+18} & 34.6 & 54.9 & 36.6 & 16.6 & 36.5 & 52.0 \\ 
\midrule
Mask RCNN-50~\cite{he2017mask} & 34.0 & 55.5 & 36.1 & 14.4 & 36.7 & 51.9 \\
Mask RCNN-101~\cite{he2017mask} & 37.5 & 60.6 & 39.9 & 17.7 & 41.0 & 55.4 \\
\bottomrule
\end{tabular}
\caption{Instance segmentation evaluation on COCO val2017. The results are shown in Mask AP.}
\label{table:seg}
\end{center}
\vspace{-0.5em}
\end{table}
\setlength{\tabcolsep}{1.4pt}

\begin{table*}
     \begin{center}
     \begin{tabular}{cccc}
     Extreme point heatmap & Center heatmap & Octagon mask & Extreme points+DEXTR~\cite{Man+18} \\
     \includegraphics[width=0.24\textwidth]{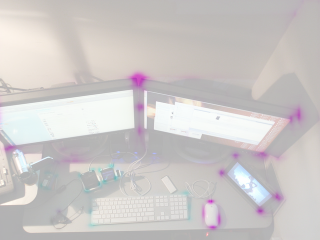}
      &\includegraphics[width=0.24\textwidth]{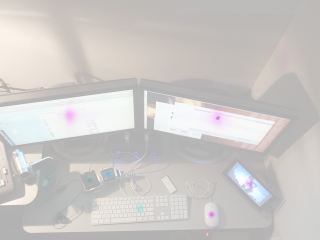}
      &\includegraphics[width=0.24\textwidth]{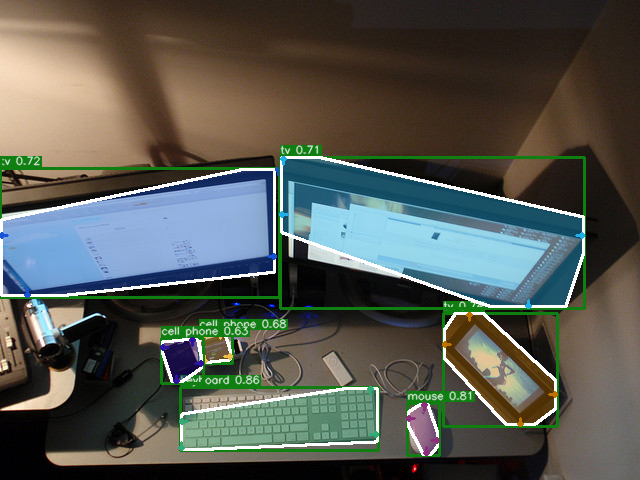}
      &\includegraphics[width=0.24\textwidth]{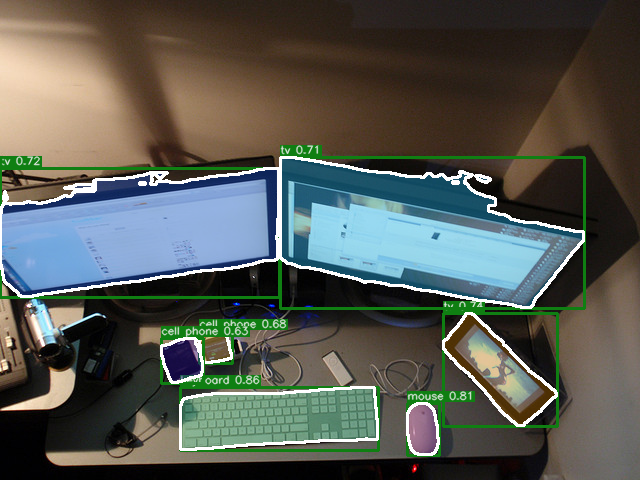} \\
     \includegraphics[width=0.24\textwidth]{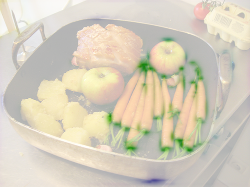}
      &\includegraphics[width=0.24\textwidth]{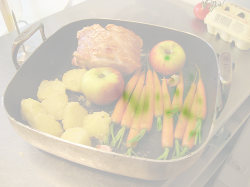}
      &\includegraphics[width=0.24\textwidth]{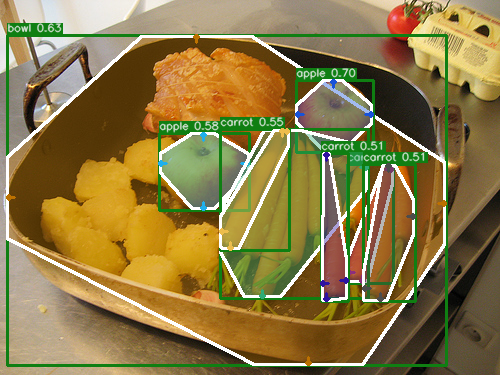}
      &\includegraphics[width=0.24\textwidth]{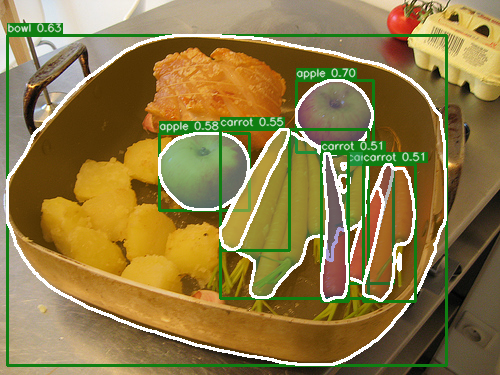} \\
     \includegraphics[width=0.24\textwidth]{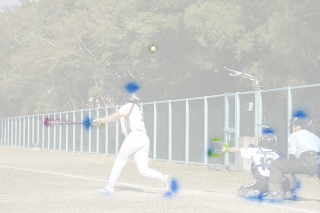}
      &\includegraphics[width=0.24\textwidth]{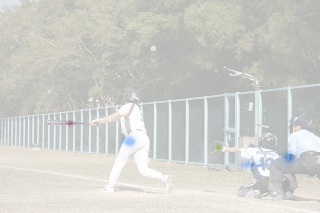}
      &\includegraphics[width=0.24\textwidth]{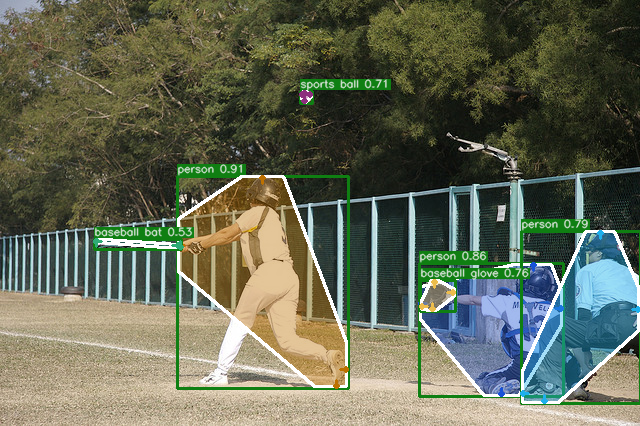}
      &\includegraphics[width=0.24\textwidth]{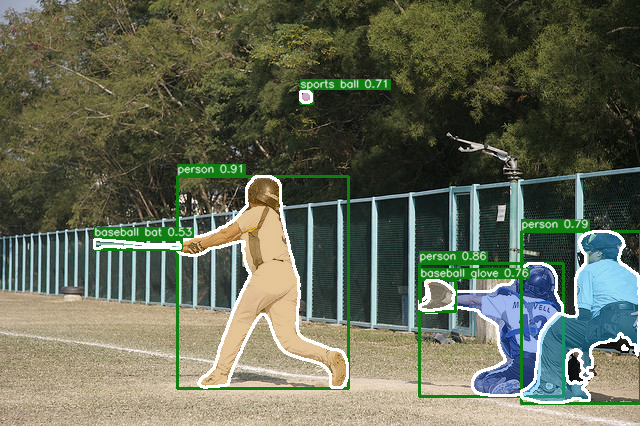} \\
     \includegraphics[width=0.24\textwidth]{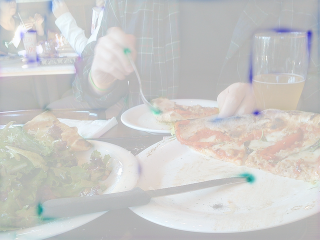}
      &\includegraphics[width=0.24\textwidth]{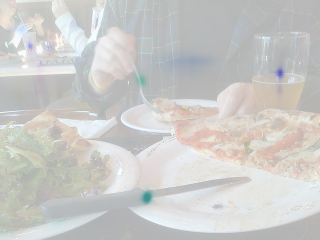}
      &\includegraphics[width=0.24\textwidth]{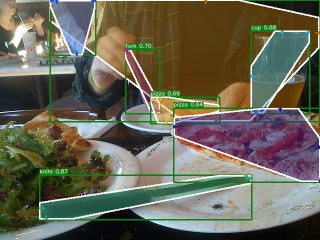}
      &\includegraphics[width=0.24\textwidth]{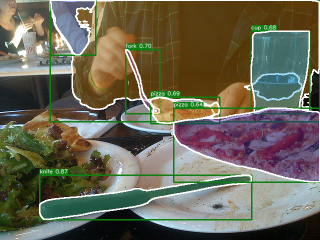} \\
      \end{tabular}
      \caption{Qualitative results on COCO val2017. First and second column: our predicted (combined four) extreme point heatmap and center heatmap, respectively. We show them overlaid on the input image. We show heatmaps of different categories in different colors. Third column: our predicted bounding box and the octagon mask formed by extreme points. Fourth column: resulting masks of feeding our extreme point predictions to DEXTR~\cite{Man+18}.}
      \label{table:demo}
      \end{center}
\end{table*}

\paragraph{Center Grouping vs. Associative Embedding}
Our ExtremeNet can also be trained with an Associative Embedding~\cite{newell2017associative} similar to CornerNet~\cite{Law_2018_ECCV}, instead of our geometric center point grouping.
We tried this idea and replaced the center map with a four-channel associative embedding feature map trained with a Hinge Loss~\cite{Law_2018_ECCV}.
Table~\ref{table:Ablation} shows the result.
We observe a $2.1\%$ AP drop when using the associative embedding.
While associative embeddings work well for human pose estimation and CornerNet, our extreme points lie on the very side of objects.
Learning the identity and appearance of entire objects from the vantage point of its extreme points might simply be too hard.
While it might work well for small objects, where the entire object easily fits into the effective receptive field of a keypoint, it fails for medium and large objects as shown in Table~\ref{table:Ablation}.
Furthermore, extreme points often lie at the intersection between overlapping objects, which further confuses the identity feature.
Our geometric grouping method gracefully deals with these issues, as it only needs to reason about appearance.

\paragraph{Edge aggregation}
Edge aggregation (Section~\ref{sec:edgeAggr}) gives a decent AP improvement of $0.7\%$.
It proofs more effective for larger objects, that are more likely to have a long axis aligned edges without a single well defined extreme point.
Removing edge aggregation improves the decoding time to 76ms and overall speed to 4.1 FPS.

\paragraph{Ghost box suppression}
Our simple ghost bounding box suppression (Section~\ref{sec:ghost}) yields $0.3\%$ AP improvement.
This suggests that ghost boxes are not a significant practical issue in MS COCO.
A more sophisticated false-positive removal algorithm, e.g., learn NMS~\cite{hosang2017learning}, might yield a slightly better result.

\paragraph{Error Analysis}
To better understand where the error comes from and how well each of our components is trained,
we provide error analysis by replacing each output component with its ground truth.
Table~\ref{table:Error} shows the result.
A ground truth center heatmap alone does not increase AP much. 
This indicates that our center heatmap is trained quite well, 
and shows that the implicit object center is learnable.
Replacing the extreme point heatmap with ground truth gives $16.3\%$ AP improvement.
When replacing both extreme point heatmap and center heatmap, the result comes to $79.8\%$, much higher than replacing one of them.
This is due to that our center grouping is very strict in the keypoint location and a high performance requires to improve both extreme point heatmap and center heatmap.
Adding the ground truth offsets further increases the AP to $86.0\%$. The rest error is from the ghost box (Section~\ref{sec:ghost}).

\subsection{State-of-the-art comparisons}

Table~\ref{table:main} compares ExtremeNet to other state-of-the-art methods on COCO test-dev.
Our model with multi-scale testing achieves an AP of $43.7$, outperforming all reported one-stage object detectors and on-par with popular two-stage detectors.
Notable, it performs $1.6\%$ higher than CornerNet, which shows the advantage of detecting extreme and center points over detecting corners with associative features. 
In single scale setting, our performance is $0.3\%$ AP below CornerNet~\cite{Law_2018_ECCV}.
However, our method has higher AP for small and median objects than CornerNet, which is known to be more challenging.
For larger objects our center response map might not be accurate enough to perform well, as a few pixel shift might make the difference between a detection and a false-negative.
Further, note that we used the half number of GPUs to train our model.

\subsection{Instance Segmentation}
Finally, we compare our instance segmentation results with/ without DEXTR~\cite{Man+18} to other baselines in Table~\ref{table:seg}.

As a dummy baseline, we directly assign all pixels inside the rectangular bounding box as the segmentation mask. 
The result on our best-model (with $43.3\%$ bounding box AP) is $12.1\%$ Mask AP.
The simple octagon mask (Section. ~\ref{sec:ins}) based on our predicted extreme points gets a mask AP of $18.9\%$, much better than the bounding box baseline.
This shows that this simple octagon mask can give a relatively reasonable object mask without additional cost.
Note that directly using the quadrangle of the four extreme points yields a too-small mask, with a lower IoU.

When combined with DEXTR~\cite{Man+18}, our method achieves a mask AP of $34.6\%$ on COCO val2017. 
To put this result in a context, the state-of-the-art Mask RCNN~\cite{he2017mask} gets a mask AP of $37.5\%$ with ResNeXt-101-FPN~\cite{lin2017feature,xie2017aggregated} backbone and $34.0\%$ AP with Res50-FPN. 
Considering the fact that our model has not been trained on the COCO segmentation annotation, or any class specific segmentations at all, our result which is on-par with Res50~\cite{he2016deep} and $2.9\%$ AP below ResNeXt-101 is very competitive. 

\section{Conclusion}
In conclusion, we present a novel object detection framework based on bottom-up extreme points estimation.
Our framework extracts four extreme points and groups them in a purely geometric manner.
The presented framework yields state-of-the-art detection results and produces competitive instance segmentation results on MSCOCO, without seeing any COCO training instance segmentations.

{
\paragraph{Acknowledgement}
We thank Chao-Yuan Wu, Dian Chen, and Chia-Wen Cheng for helpful feedback.
}

{\small
\bibliographystyle{ieee}
\bibliography{egbib}
}

\end{document}